\documentclass[runningheads]{llncs}
\usepackage{graphicx}

\usepackage{tikz}
\usepackage{comment}
\usepackage{amsmath,amssymb} 
\usepackage{color}
\usepackage{orcidlink}

\usepackage{hyperref}
\newcommand\nnfootnote[1]{%
  \begin{NoHyper}
  \renewcommand\thefootnote{}\footnote{#1}%
  \addtocounter{footnote}{-1}%
  \end{NoHyper}
}

\usepackage{enumitem,xspace}
\usepackage{algpseudocode}
\usepackage{algorithm}
\usepackage{multirow}
\usepackage{booktabs}

\usepackage[accsupp]{axessibility}  

\begin{document}
\pagestyle{headings}
\mainmatter
\def\ECCVSubNumber{8106}  

\title{Bitwidth-Adaptive Quantization-Aware Neural Network Training: A Meta-Learning Approach} 

\titlerunning{Bitwidth-Adaptive QAT: A Meta-Learning Approach}
%
\author{Jiseok Youn\inst{1}\orcidlink{0000-0003-1347-8874}\index{Youn, Jiseok} \and
Jaehun Song\inst{2}\orcidlink{0000-0001-8573-3331} \and
Hyung-Sin Kim\inst{2}*\orcidlink{0000-0001-8605-5077} \and
Saewoong Bahk\inst{1}*\orcidlink{0000-0002-4771-3927}}
\authorrunning{J. Youn et al.}
%
\institute{Department of Electrical and Computer Engineering and INMC, \\ Seoul National University, Seoul, South Korea \\ \email{jsyoun@netlab.snu.ac.kr, sbahk@snu.ac.kr} \and
Graduate School of Data Science, Seoul National University, Seoul, South Korea \\ \email{\{steve2972, hyungkim\}@snu.ac.kr}}

\maketitle

\newcommand{\proposal}{MEBQAT\xspace}

\nnfootnote{*Corresponding authors}

\begin{abstract}
Deep neural network quantization with \textit{adaptive bitwidths} has gained increasing attention due to the ease of model deployment on various platforms with different resource budgets. In this paper, we propose a meta-learning approach to achieve this goal. Specifically, we propose \proposal, a simple yet effective way of bitwidth-adaptive quantization-aware training (QAT) where meta-learning is effectively combined with QAT by redefining meta-learning tasks to incorporate bitwidths. 
After being deployed on a platform, \proposal allows the (meta-)trained model to be quantized to any candidate bitwidth 
with minimal inference accuracy drop. Moreover, in a few-shot learning scenario, \proposal can also adapt a model to any bitwidth as well as any \textit{unseen} target classes by adding conventional optimization or metric-based meta-learning.

We design variants of \proposal to support both (1) a bitwidth-adaptive quantization scenario and (2) a new few-shot learning scenario where both quantization bitwidths and target classes are jointly adapted. 
Our experiments show that merging bitwidths into meta-learning tasks results in remarkable performance improvement: 98.7\% less storage cost compared to bitwidth-dedicated QAT and 94.7\% less back propagation  compared to bitwidth-adaptive QAT in bitwidth-only adaptation  scenarios, while improving classification accuracy by up to 63.6\% compared to vanilla meta-learning in bitwidth-class joint adaptation scenarios.
\end{abstract}

\section{Introduction}
Recent development in deep learning has provided key techniques for equipping resource-constrained devices with larger networks by reducing neural network computational costs. To this end, several research directions have emerged such as network optimization
\cite{LiangGWSZ21,MathieuHL13}, 
parameter factorization
\cite{SainathKSAR13,SwaminathanGKA20}, 
network pruning
\cite{SuzukiAMHIWHYN20,YuLCLMHGLD18}, 
and quantization \cite{ChoukrounKYK19,JacobKCZTHAK18,ZhangYYH18,ZhouNZWWZ16}.
In particular, quantization can significantly reduce model size, computational requirements and power consumption by expressing model weights and activations in lower precision. 
For example, quantizing a model from FP32 to Int8 with devices equipped with fast arithmetic hardware units for low-precision operands can reduce inference delay by up to 5$\times$~\cite{nvidia_quantization}. 

However, one challenge associated with quantization is the difficulty of tailoring models to various bitwidths to compensate for platforms with different resource constraints. 
This is especially important in situations where a quantized model is deployed to platforms with different battery conditions, hardware limitations, or software versions. 
In order to solve this problem, a recent trend in quantization gave rise to adaptive bitwidths, which allows models to \textit{adapt} to bitwidths of varying precision \cite{BulatT21,JinYL20,SunLRHLJL21}.

\begin{figure}[t]
\centering
\centerline{\includegraphics[width=\textwidth]{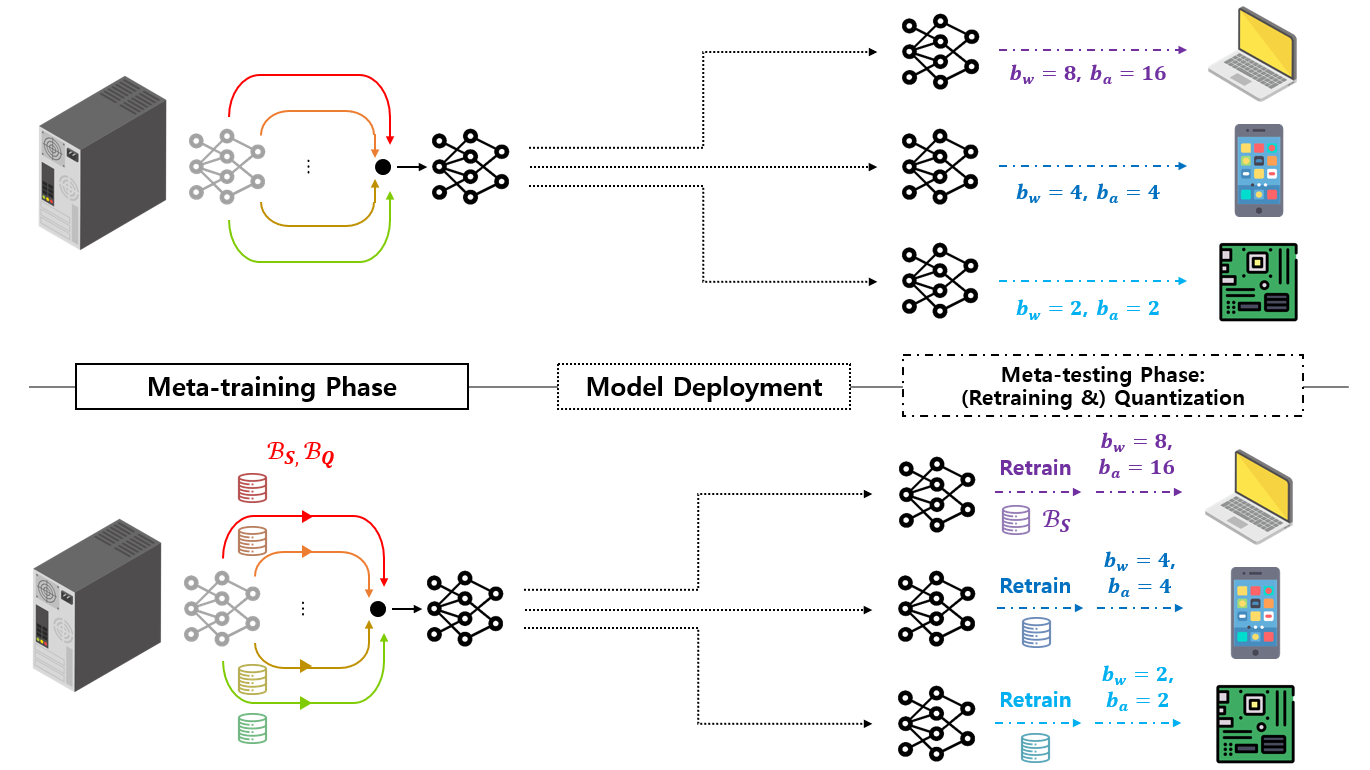}}
\caption{Overview of \proposal on bitwidth-only adaptation (above) and bitwidth-class joint adaptation (below) scenarios.}
\label{fig:mebqat_overview}
\end{figure}

In this paper, we provide a different perspective on this research direction by considering a modified formulation of bitwidth-adaptive quantization-aware training (QAT) with  \textit{meta-learning}~\cite{FinnAL17}, as shown in Figure~\ref{fig:mebqat_overview}.
In typical meta-learning scenarios, a meta task is defined as a subset of training data, divided on the basis of class~\cite{FinnAL17} or data configuration~\cite{gong2019metasense,cho2021camera}. With this task definition, the meta-training phase requires a tailored, large-scale dataset for a model to experience many meta tasks while the meta-testing phase needs the model to be retrained with few-shot data for a target task. 
To apply meta-learning in bitwidth-adaptive QAT, we propose \proposal by newly defining a meta task to incorporate a \textit{bitwidth setting}, a model hyperparameter independent of the dataset.
Thus, our meta task definition enables \textit{dataset-agnostic} meta-learning: meta-learning without the need for few-shot-learning-specific datasets. In the meta-testing phase, the model is not retrained but quantized immediately with any target bidwidth, resulting in fast adaptation. Experiments show that \proposal performs comparably to state-of-the-art bitwidth-adaptive QAT schemes. The results suggest that bitwidth-adaptive QAT can be categorized as a meta learning problem.

In addition, to show that \proposal is synergistically combined with typical meta-learning scenarios, we also investigate a new meta-learning scenario where quantization bitwidths and target classes are \textit{jointly adapted}. In this scenario, we define a meta task as a combination of bitwidth setting and target classes. With these modified tasks, we show that \proposal can be merged with both optimization-based meta-learning (Model-Agnostic Meta-Learning (MAML)~\cite{FinnAL17}) and metric-based meta-learning (Prototypical Networks~\cite{SnellSZ17}) frameworks. 
Experiments show that \proposal produces a model that is not only adaptable to arbitrary bitwidth settings, but also robust to unseen classes when retrained with few-shot data in the meta-testing phase. \proposal significantly outperforms both vanilla meta learning and the combination of dedicated QAT and meta learning, demonstrating that \proposal successfully merges bitwidth-adaptive QAT and meta learning without losing their own advantages. With this new scenario given by \proposal, a model can be deployed on more various platforms regardless of their resource constraints and classification tasks.

To summarize, our contributions are three-fold:
\begin{itemize}
\item We propose \textbf{ME}ta-learning based \textbf{B}itwidth-adaptive \textbf{QAT} (\proposal), by newly defining meta-learning tasks to include bitwidth settings and averaging gradients approximated for different bitwidths over those tasks to incorporate the essence of quantization-awareness.
\item We show that our method can obtain a model robust to various bitwidth settings by conducting extensive experiments on various supervised-learning contexts, datasets, model architectures, and QAT schemes. In the traditional classification problem, \proposal shows comparable performance to existing bitwidth-adaptive QAT methods and dedicated training of the model to a given bitwidth, but with higher training efficiency (94.7\% less back-propagations required).
\item We define a new few-shot classification context for \proposal where both bitwidths and classes are jointly adapted using few-shot data. \proposal well fits both optimization- and metric-based meta-learning frameworks.
In terms of classification accuracy, \proposal outperforms vanilla meta-learning by up to 63.6\% and a na\"ive combination of dedicated QAT and meta learning by up to 27.48\%, while also adding bitwidth adaptability comparable to state-of-the-art bitwidth-adaptive QAT methods.
\end{itemize}

\section{Related Work}
\subsubsection{Quantization-Aware Training.}
Existing approaches to quantization can be broadly split into Post-Training Quantization (PTQ) and QAT. PTQ quantizes a model trained without considering quantization, and requires sophisticated methods such as solving optimization problems~\cite{BaiCHS21,ChoukrounKYK19,FangSATGH20,CaiYDGMK20} and model reconstruction~\cite{HubaraNHBS21,LiGTYHZYWG21}. However, given that most platforms that utilize a quantized model are resource constrained, these methods can incur significant computational burden.
To reduce post-training computation, we instead focus on QAT, which trains a model to alleviate the drop of accuracy when quantized. QAT methods usually define a formula for approximating a gradient of a quantization function output w.r.t. the input. To this end, recent works suggest integer-arithmetic-only quantization methods \cite{JacobKCZTHAK18,ZhangYYH18,JungSLSHKHC19,EsserMBAM20} and introduce differentiable asymptotic functions for non-differentiable quantization functions \cite{GongLJLHLYY19}.

However, conventional QAT is limited in that a model is trained for a single dedicated bitwidth, showing significant performance degradation when quantized to other bitwidths. In other words, supporting multiple bitwidths would require training multiple copies of the model to each bitwidth.

\subsubsection{Bitwidth-adaptive QAT.}
In order to overcome such shortcomings, some QAT-based approaches aim to train a model only once and use it on various bitwidth settings. A number of studies~\cite{WuWZTVK18,WangLLLH19,GuoZMHLWS20,CuiLY0CKX20,WangWCLLWLH20,ShenLLLSYO20,BaiCHS21} use Neural Architecture Search (NAS) to train a \textit{super-network} involving multiple bitwidths in a predefined search space and sample a sub-network quantized with the target bitwidth setting given or searched taking the hardware into account. However,  NAS-based approaches usually suffer from difficult training, heavy computation, and collapse on 8-bit precision without special treatment.

AdaBits \cite{JinYL20} was the first to propose another research direction, namely the concept of training a single model adaptive to any bitwidth. Specifically, the model is trained via joint quantization and switchable clipping level. As a similar approach, Any-precision DNN \cite{YuLSH021} enables adaptable bitwidths via knowledge distillation \cite{HintonVD15} and switchable Batch Normalization (BN) layers. The authors in \cite{SunLRHLJL21} utilize wavelet decomposition and reconstruction \cite{Mallat89} for easy bitwidth adjustment by adjusting hyperparameters. Furthermore, Bit-mixer \cite{BulatT21} aims to train a mixed-precision model where its individual layers can be quantized to an arbitrary bitwidth. Although these methods allow a single model to train for multiple bitwidths, some parts of the model (e.g., BN layers) still need to be trained dedicated to each precision candidate which increases the number of parameters w.r.t. the number of bitwidth candidates \cite{CuiLY0CKX20}. Moreover, prior work solely focuses on model quantization and ignores the possibility that users require slightly different tasks that the pretrained model does not support.

\subsubsection{Meta learning.}
Meta learning has recently attracted much attention in the research community due to its potential to train a model that can flexibly adapt to different tasks, even with a few gradient steps and limited amounts of labeled data, making it ideal for resource-constrained platforms \cite{FinnAL17,SungZXHY17}. 

One of the most common approaches to meta learning is optimization-based meta learning that trains a base model from which a model starts to be adapted to a given task by using experience from many different tasks. Model-Agnostic Meta Learning (MAML) \cite{FinnAL17} suggests to learn from multiple tasks individually, evaluate the overall adaptation performance, and learn to increase it. Many variants of MAML have emerged to improve upon this method\cite{LiZCL17,FinnXL18,YoonKDKBA18,RaviB19,RajeswaranFKL19,TriantafillouZDLEXGGSML20,ZhouZYFXH21}. 
Another approach to meta learning is metric-based meta learning, which attempts to learn an embedding function such that an unseen class can be predicted by seeking the label with minimum distance. 
Prototypical Networks \cite{SnellSZ17} calculates prototypes as a milestone for each label by averaging the corresponding embeddings.
While meta learning provides a personalizable model robust to unseen classes, there is still a lack of research concerning the applicability of meta learning in quantization.

To the best of our knowledge, this work is the first to show that bitwidth-adaptive QAT and meta learning can be merged synergistically without sacrificing their own advantages. Specifically, our proposal \proposal provides bitwidth-adaptive QAT with zero-copies of any part of the model. In addition, by defining a \textit{meta task} as a combined set of bitwidth setting and target classes, \proposal produces a model that quickly adapts to arbitrary bitwidths as well as target classes.

\section{Meta-Learning Based Bitwidth-Adaptive QAT}
\begin{figure}[t]
\centering
\centerline{\includegraphics[width=\textwidth]{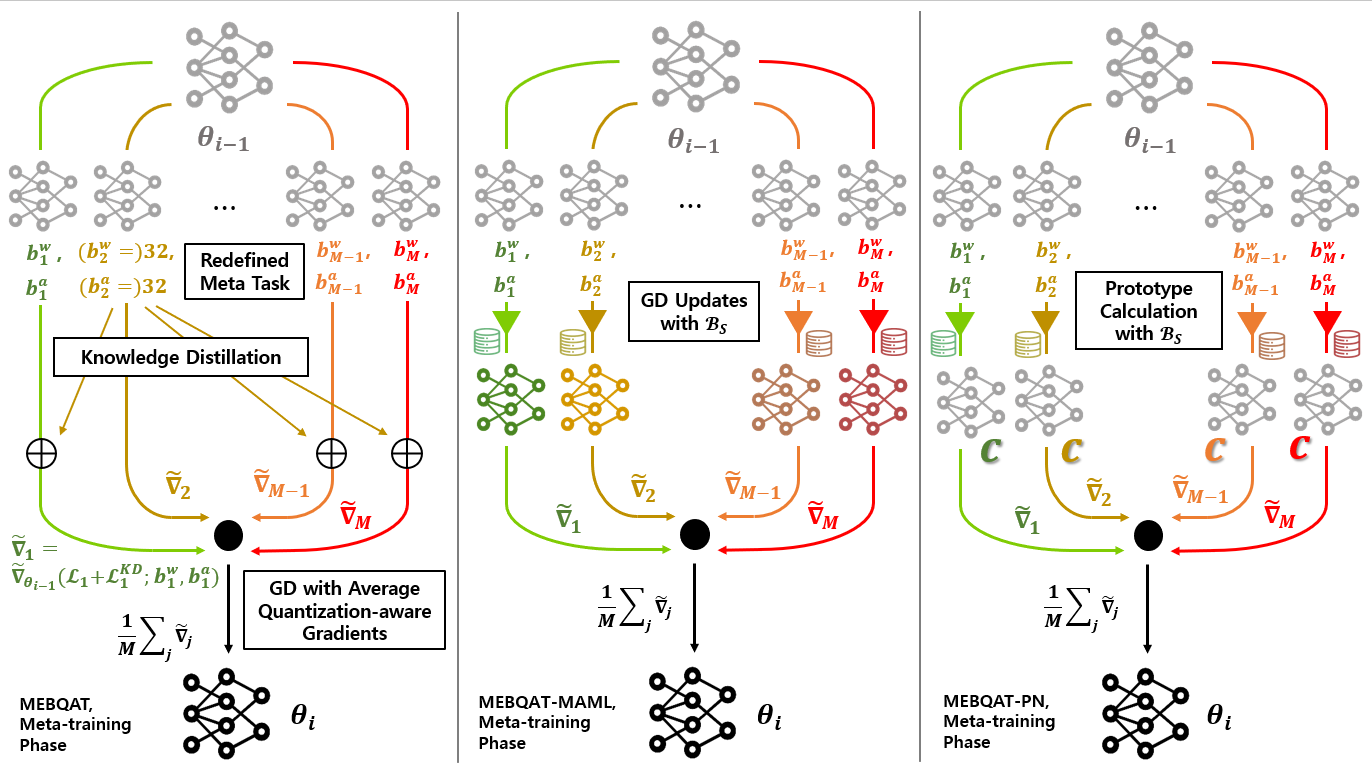}}
\caption{Illustration of meta-training phase in \proposal, \proposal-MAML, and \proposal-PN. GD stands for Gradient Descent. $\mathbf{c}$ denotes prototypes. $\mathcal{B}_S$ represents a support set. Illustration of adaptation and inference phase is provided in appendix.}
\label{fig:mebqat_phase1_variants}
\end{figure}

In this section, we introduce \textbf{ME}ta-learning based \textbf{B}itwidth-adaptive \textbf{QAT} (\proposal), a once-for-all method that aims to provide a model adaptable to any bitwidth setting by synergistically combining QAT with meta-learning methodologies. 
Similar to conventional meta-learning schemes, \proposal operates in two phases: a meta-training and a meta-testing phase. In the meta-training phase, \proposal trains a base model by experiencing various tasks to improve its adaptability. Importantly, the meta-training phase performs QAT with the task definition including bitwidth settings to support bitwidth-adaptive QAT. In the meta-testing phase, the meta-trained base model is deployed at a platform and tailored for a platform-specific target task.

We consider two practical scenarios for \proposal, (1) bitwidth adaptation scenario and (2) bitwidth-class joint adaptation scenario. The former scenario is the main problem that bitwidth-adaptive QAT methods target, and as such, we aim to provide similar performance to state-of-the-art schemes but with less complexity, using our meta-learning-based approach. The bitwidth-class joint adaptation scenario is a new scenario in which a model can adapt to not only an arbitrary target bitwidth but also unseen target classes. To support these scenarios, we provide three variants of \proposal, called MEBQAT, -MAML, and -PN, as shown in Figure~\ref{fig:mebqat_phase1_variants}. Following convention in \cite{FinnAL17}, MEBQAT, -MAML, -PN aim to optimize Equations \ref{eq:optim_goal1}-\ref{eq:optim_goal3}, respectively.

\begin{equation}
\label{eq:optim_goal1}
{\min}_{\theta}{\sum}_{j}{\mathcal{L}}_{j}={\sum}_{j}{\mathcal{L}}(f_{\text{Quantize}(\theta; b^w_j,b^a_j)})
\end{equation}
\begin{equation}
\label{eq:optim_goal2}
{\min}_{\theta}{\sum}_{j}{\mathcal{L}}^{Q}_{j}={\sum}_{j}{\mathcal{L}}^{Q}(f_{\text{Quantize}(\{{\theta}-{\alpha}{\tilde{\nabla}}_{\theta}{\mathcal{L}^{S}}(f_{\theta};b^w_j,b^a_j)\}; b^w_j,b^a_j)})
\end{equation}
\begin{equation}
\label{eq:optim_goal3}
{\min}_{\theta}{\sum}_{j}{\mathcal{L}}^{Q}_{j}={\sum}_{j}{\mathcal{L}}^{Q}(f_{\text{Quantize}(\theta; b^w_j,b^a_j)}, \boldsymbol{c}^{S}(f_{\text{Quantize}(\theta; b^w_j,b^a_j)}))
\end{equation}

\begin{table}[t]
\caption{Summary of notations.}
\centering
\resizebox{0.9 \columnwidth}{!}{%
\label{table:fig1and2_notation}
\begin{tabular}{l|l}
\hline
Notation in Fig. 1 & Meaning \\ \hline
$\mathcal{B}_S / \mathcal{B}_Q$ &  Support data for adaptation / Query data for inference \\
$b_w / b_a$                     &  Test bitwidth of weights / activations \\ \hline
Notation in Fig. 2 & Meaning \\ \hline
$\mathcal{B}_S / \mathcal{B}_Q$ &  Support data for adaptation / Query data for inference \\
${\theta}_{i}$                  &  Model parameters after $i$-th optimization or update \\
$M$                             &  Number of inner-loop (task)s per outer-loop (i.e., meta batch size) \\
$b_j^w / b_j^a$                 &  Training bitwidth of weights / activations in $j$-th meta-task $(j=1,2,\cdots,M)$ \\
$\boldsymbol{c}$                &  Prototype in PN framework, differentiated by colors \\
${\mathcal{L}}_{j}$             &  Loss in $j$-th meta-task \\
${\mathcal{L}}_{j'}^{KD}$       &  Distillation loss in $j'$-th meta-task $(j'=2,\cdots,M)$\\
${\tilde{\nabla}}_{j}$          &  Gradients in $j$-th meta-task, approximated according to the bitwidth $(b_j^w, b_j^a)$  \\ \hline
\end{tabular}}
\end{table}

\subsection{Bitwidth Adaptation Scenario}

In this scenario, we assume that users' target classes are the same as those used for model training; in other words, both meta-training and meta-testing phases have the same classification task. However, each user may have different target bitwidths considering its own resource budget. Therefore in the meta-testing phase, a user immediately quantizes the base model using its own bitwidth setting without the need for fine-tuning. 

\begin{algorithm}[t]
   \caption{\proposal, meta-training phase}
   \label{algo:mebqat-1_phase1}
\begin{algorithmic}
   \State Initialize base model parameters $\theta_0$, bitwidth task set $\mathcal{T}_b$, training set $\mathcal{D}$ comprising $(\mathbf{x},y)$, and step size $\beta$
   \For{epoch $i=1$ {\bfseries to} $E$}
       \State $\mathcal{B} \gets$ random sample from $\mathcal{D}$
       \State $\hat{y}_{\mathbf{x}} \gets f_{\theta_{i-1}}(\mathbf{x})$ for all $\mathbf{x} \in \mathcal{B}$ \Comment{Get soft labels using full precision} 
       \For{$j=1$ {\bfseries to} $M$}
           \State $(b^w_j,b^a_j) \gets$ random sample from $\mathcal{T}_b$ \Comment{Sample a bitwidth task} 
           \State $\phi \gets \text{Quantize}(\theta_{i-1}; b^w_j,b^a_j)$ 
           \State $\mathcal{L}_{j} \gets \frac{1}{|\mathcal{B}|}\sum_{(\mathbf{x}, y){\in}\mathcal{B}}\mathcal{L}(f_{\phi}(\mathbf{x}), y)$ \Comment{Get task-specific supervised loss}
           \State $\mathcal{L}_{j}^{KD} \gets \frac{1}{|\mathcal{B}|}\sum_{(\mathbf{x}, y){\in}\mathcal{B}}\mathcal{L}(f_{\phi}(\mathbf{x}), \hat{y}_{\mathbf{x}})$ \Comment{Get task-specific KD loss}
           \State $\tilde{\nabla}_{j} \gets {\tilde{\nabla}}_{\theta_{i-1}}(\mathcal{L}_{j}+\mathcal{L}_{j}^{KD} ;b_{j}^{w}, b_{j}^{a})$ \Comment{Get task-specific, quant-aware gradient}
       \EndFor
       \State $\theta_i \gets \theta_{i-1}-\frac{\beta}{M}\sum_{j=1}^{M}{\tilde{\nabla}}_{j}$ \Comment{Update base model}
   \EndFor
\end{algorithmic} 
\end{algorithm}

To support this scenario, we define a bitwidth task set $\mathcal{T}_b$ that consists of various tuples $(b^w, b^a)$ where $b^w$ and $b^a$ are bitwidths for weight quantization and activation quantization, respectively. Assuming that $f_\theta$ is a base model parameterized by $\theta$, \proposal aims to meta-train $\theta$ by experiencing many bitwidth tasks.
In contrast to typical meta-learning scenarios, task-specific data samples are not needed because data is decoupled from task definition. Therefore, in each epoch $i$, \proposal samples a (common) batch of data $\mathcal{B}$ from training set $\mathcal{D}$ that all of $M$ sampled tasks share. 
In addition, before entering into task-specific operation, \proposal gets soft labels $\hat{y}_\mathbf{x}$ for the batch $\mathcal{B}$ using the current full precision model parameters $\theta_{i-1}$ to utilize knowledge distillation~\cite{HintonVD15} as in Any-precision DNN~\cite{YuLSH021}. The idea is that the full precision model has more information and can teach a quantized model.

\proposal samples $M$ bitwidth tasks in each epoch $i$. For each selected bitwidth tuple $(b^w_j,b^a_j)$, the full precision model $\theta_{i-1}$ is quantized to $\phi$ using the tuple $(b^w_j,b^a_j)$ and two types of task-specific losses are calculated based on the quantized model $\phi$: supervised loss $\mathcal{L}_j$ and knowledge distillation loss $\mathcal{L}_j^{KD}$. Note that when $(b^w_j,b^a_j)$ happens to be (FP,FP) (full precision), $\mathcal{L}_j^{KD}$ becomes zero. 
Using the loss $\mathcal{L}_j+\mathcal{L}_j^{KD}$, task-specific gradients $\tilde{\nabla}_{j}$ are calculated in a quantization-aware manner. Quantization-aware gradient calculation considers the sample bitwidth $(b^w_j,b^a_j)$ and detailed method depends on which QAT scheme is combined with \proposal. Lastly, model parameters are updated to $\theta_i$ using gradient descent with step size $\beta$·
Algorithm~\ref{algo:mebqat-1_phase1} illustrates this process.

\subsection{Bitwidth-Class Joint Adaptation Scenario}

In this section, we propose a new meta-learning scenario for bitwidth-class joint adaptation where users may have their own target bitwidths and classification tasks. Each user is assumed to have a small local dataset for their classification tasks, which is used in the meta-testing phase to retrain the base model. Specifically, we consider $N$-way $K$-shot tasks where $N$ is the number of target classes and $K$ is the number of data samples per each of $N$ classes. Assuming that $\mathcal{Y}$ is a set of randomly selected $N$ classes, a single joint task including both bitwidths and classes is defined as $(b_w, b_a, \mathcal{Y})$.

To support this new scenario, we design two types of \proposal: (1) \proposal-MAML, which adopts a representative optimization-based meta-learning framework MAML~\cite{FinnAL17} and (2) \proposal-PN, which adopts a representative metric-based meta-learning framework called Prototypical Networks (PN)~\cite{SnellSZ17}.

\begin{algorithm}[t]
   \caption{\proposal-MAML, meta-training phase}
   \label{algo:mebqat-2_phase1}
\begin{algorithmic}
   \State Initialize base model parameters $\theta_0$, bitwidth  task set $\mathcal{T}_b$, training set $\mathcal{D}$ comprising $(\mathbf{x},y)$, and step sizes $\alpha$, $\beta$.
   \For{epoch $i=1$ {\bfseries to} $E$}
        \For{$j=1$ {\bfseries to} $M$} 
           \State $(b^w_j,b^a_j) \gets$ random sample from $\mathcal{T}_b$ \Comment{Sample a bitwidth task}
            \State $\mathcal{Y}_j \gets$ a set of randomly selected $N$ classes \Comment{Sample a data task}
        \State $\mathcal{D}_{\mathcal{Y}_j} \gets$ Subset of $\mathcal{D}$ where $y \in \mathcal{Y}_j$
   
           \State $\mathcal{B}_{S} \gets$ random sample from $\mathcal{D}_{\mathcal{Y}_j}$ \Comment{$N$-way $K$-shot support set}
           \State $\phi_0 \gets \theta_{i-1}$
           \For{$u=1$ {\bfseries to} $U$} \Comment{Gradient decent with quantization}
               \State $\phi_{u-1}^q \gets \text{Quantize}(\phi_{u-1}; b^w_j, b^a_j)$
               \State $\mathcal{L}_{j}^{S} \gets \frac{1}{|\mathcal{B}_S|}\sum_{(\mathbf{x}, y){\sim}\mathcal{B}_S}\mathcal{L}(f_{\phi_{u-1}^q}(\mathbf{x}), y)$
               \State ${\phi}_{u} \gets {\phi}_{u-1}-{\alpha}\tilde{\nabla}_{{\phi}_{u-1}}(\mathcal{L}_{j}^{S};b_{j}^{w}, b_{j}^{a})$
           \EndFor
            \State $\phi_{U}^q \gets \text{Quantize}(\phi_{U}; b^w_j, b^a_j)$
           \State $\mathcal{B}_{Q} \gets$ random sample from $\mathcal{D}_{\mathcal{Y}_j} \setminus \mathcal{B}_S$ \Comment{$N$-way $K$-shot query set}
           \State $\mathcal{L}_{j}^{Q} \gets \frac{1}{|\mathcal{B}_Q|}\sum_{(\mathbf{x}, y){\sim}\mathcal{B}_Q}\mathcal{L}(f_{{\phi}_U^q}(\mathbf{x}), y)$
           \State $\tilde{\nabla}_{j} \gets {\tilde{\nabla}}_{\theta_{i-1}}(\mathcal{L}_{i}^Q; b_{j}^{w}, b_{j}^{a})$ \Comment{Get task-specific, quant-aware gradient}
       \EndFor
       \State $\theta_i \gets  \theta_{i-1}-\frac{\beta}{M}\sum_{j=1}^{M}{\tilde{\nabla}}_{j}$ \Comment{Update base model}
   \EndFor 
\end{algorithmic}
\end{algorithm}

\subsubsection{\proposal-MAML.}

The main difference between \proposal-MAML and \proposal lies in the inner-loop operation for each task. In each iteration $j$ of the inner loop, \proposal-MAML samples both a bitwidth task $(b^w_j,b^a_j)$ and a data task $\mathcal{Y}_j$. Note that the bitwidth task is newly added to the original MAML operation.
Assuming that the current model in epoch $i$ is $\theta_{i-1}$, the model is updated to a task-specific quantized model $\phi^q_U$ by using $U$-step gradient decent and a QAT method with the bitwidth setting $(b^w_j,b^a_j)$ and a task-specific support set $\mathcal{B}_S$. Given the task-specific model $\phi^q_U$ and a query set $\mathcal{B}_Q$, task-specific loss $\mathcal{L}^Q_j$ and gradient $\tilde{\nabla}_j$ are calculated in a quantization-aware manner. Given that $\tilde{\nabla}_j$ requires second-order gradient calculation which is computationally expensive, we instead adopt a first-order approximation of MAML, called FOMAML. Algorithm \ref{algo:mebqat-2_phase1} illustrates the process.

In the meta-testing phase, a user retrains the base model using a local support set of its own classification task and quantizes the model using its own bitwidth setting. Then the model performance is evaluated by inferencing data points in a local query set. Given that a user platform is likely to be resource constrained, the number of gradient decent updates in the meta-testing phase can be smaller than $U$, as in the original MAML.

\begin{algorithm}[t]
   \caption{\proposal-PN, meta-training phase}
   \label{algo:mebqat-3_phase1}  
\begin{algorithmic}
   \State Initialize base model parameters $\theta_0$, bitwidth  task set $\mathcal{T}_b$, training set $\mathcal{D}$ comprising $(\mathbf{x},y)$, and step sizes $\beta$.
   \For{epoch $i=1$ {\bfseries to} $E$}
       \State $\mathcal{Y}_i \gets$ a set of randomly selected $N$ classes \Comment{Sample a data task}
      \State $\mathcal{B}_{S} \gets$ random sample from $\mathcal{D}_{\mathcal{Y}_i}$ \Comment{$N$-way $K$-shot support set}
      \State $\mathcal{B}_{Q} \gets$ random sample from $\mathcal{D}_{\mathcal{Y}_i} \setminus \mathcal{B}_{S}$ \Comment{$N$-way $K$-shot query set}
       \For{$j=1$ {\bfseries to} $M$}
            \State $(b^w_j,b^a_j) \gets$ random sample from $\mathcal{T}_b$ \Comment{Sample a bitwidth task}
            \State $\phi \gets \text{Quantize}(\theta_{i-1}; b^w_j, b^a_j)$
           \For{$n \in \mathcal{Y}_i$} \Comment{Get prototypes using support set}
                \State $\mathbf{c}_{n} \gets \frac{1}{K}{\sum_{(\mathbf{x}, y){\in}\mathcal{B}_{S},y=n}}{f_{\phi}(\mathbf{x})}$
           \EndFor
           \State $\mathcal{L}_{j}^{Q}=0$
           \For{$n \in \mathcal{Y}_i$} \Comment{Get task-specific loss using query set}
              \For{$(\mathbf{x}, y) \in \mathcal{B}_{Q}$ where $y=n$}
                 \State $\mathcal{L}_{j}^{Q} = \mathcal{L}_{j}^{Q} + \frac{1}{NK}[d(f_{\phi}(\mathbf{x}), \mathbf{c}_{n}) + {\log}{\sum_{n'}}{\exp(-d(f_{\phi}(\mathbf{x}), \mathbf{c}_{n'}))}]$ 
              \EndFor
           \EndFor
           \State $\tilde{\nabla}_{j} \gets {\tilde{\nabla}}_{\theta_{i-1}}(\mathcal{L}_{j}^{Q};b_{j}^{w}, b_{j}^{a})$. \Comment{Get task-specific, quant-aware gradient}
       \EndFor
       \State $\theta_i \gets  \theta_{i-1}-\frac{\beta}{M}\sum_{j=1}^{M}{\tilde{\nabla}}_{j}$ \Comment{Update base model}
   \EndFor
\end{algorithmic}
\end{algorithm}

\subsubsection{\proposal-PN.} 

A limitation of \proposal-MAML arises from the necessity of gradient decent-based fine-tuning in the meta-testing phase, which can become a computational burden to resource-constrained platforms. 
In contrast to MAML, Prototypical Network (PN) trains an embedding function such that once data points are converted into embeddings, class prototypes are calculated using a support dataset and query data is classified by using distance from each class prototype. Therefore in the meta-testing phase, \proposal-PN does not require gradient descent but simply calculates class prototypes using a local support set, which significantly reduces computation overhead.

Algorithm \ref{algo:mebqat-3_phase1} illustrates \proposal-PN's meta-training phase. Unlike original PN, to include a bitwidth setting in a task in each epoch $i$, \proposal samples bitwidths $(b^w_i, b^a_i)$ as well as target classes $\mathcal{Y}_i$, quantizes the current model  $\theta_{i-1}$ to $\phi$ using the selected bitwidths, and calculates class prototypes $\mathbf{c}_n$ for  $n \in \mathcal{Y}_i$ using the quantized model $\phi$ and a support set $\mathcal{B}_S$.
Then task-specific loss $\mathcal{L}^Q_j$ is calculated using distance between embeddings for query data points in $\mathcal{B}_Q$ and class prototypes. Lastly, task-specific gradient $\tilde{\nabla}_i$ is computed in a quantization-aware manner.

\subsection{Implementation}

We also include specific implementation details to improve the training process of \proposal.
First, in each epoch of \proposal (Algorithm~\ref{algo:mebqat-1_phase1}), we fix the bitwidth task in the first inner-loop branch to full-precision (FP,FP) instead of a random sample. This implementation is required for the base model to experience full precision in every epoch, thus improving accuracy.
Second, while sampling random bitwidth settings, we exclude unrealistic settings such as (FP,1) and (1,FP) because these settings not only are impractical and improbable, but also hinder convergence.
Third, when there are some minor bitwidth settings that a QAT scheme treats differently from other bitwidths (e.g., 1-bit of DoReFa-Net \cite{ZhouNZWWZ16}), we sample the minor settings more frequently (e.g. at least once in each epoch).

\section{Evaluation}

To demonstrate the validity of \proposal, we conduct extensive experiments on multiple supervised-learning contexts, datasets, model architectures, and configurations of quantization.

\subsection{Experiments on the Bitwidth Adaptation Scenario}
\label{section:experiments_traditional}

In the bitwidth adaptive scenario with shared labels, we compare \proposal with (1) (bitwidth-dedicated) QAT and (2) existing bitwidth-adaptive QAT methods (AdaBits \cite{JinYL20} and Any-precision DNN (ApDNN) \cite{YuLSH021}).

\proposal adopts multiple quantization configurations 
depending on the compared scheme. When compared with AdaBits, \proposal quantizes a tensor and approximates its gradient using the same Scale-Adjusted Training (SAT) \cite{JinYLQ20} that AdaBits adopts, with $\mathcal{T}_{b}=\{2, 3, 4, 5, 6, 7, 8, 16, \text{FP}\}$ where FP denotes full-precision Float32.  Furthermore, just as in AdaBits, we quantize the first and last layer weights into 8-bits with BN layers remaining full-precision. When compared with Any-precision DNN, \proposal quantizes a tensor and approximates its gradient in a DoReFa-Net based manner, with $\mathcal{T}_{b}=\{1, 2, 3, 4, 5, 6, 7, 8, 16, \text{FP}\}$. Note that we differentiate the formula for 1-bit and other bitwidths as DoReFa-Net \cite{ZhouNZWWZ16} does. In this case, we do not quantize the first, last, and BN layers. The number of inner-loop tasks per task is set to 4.

Optimizer and learning rate scheduler settings depend on the model architecture and dataset used. For MobileNet-v2 on CIFAR-10, we use an Adam optimizer for 600 epochs with an initial learning rate $5{\times}{10^{-2}}$ and a cosine annealing scheduler without restart. For pre-activation ResNet-20 on CIFAR-10, we use an AdamW optimizer for 400 epochs with an initial learning rate ${10^{-3}}$ divided by 10 at epochs \{150, 250, 350\}. Finally, for the 8-layer CNN in \cite{YuLSH021} on SVHN, we use a standard Adam optimizer for 100 epochs with an initial learning rate ${10^{-3}}$ divided by 10 at epochs \{50, 75, 90\}. 

Finally, as in MAML, all BN layers are used in a transductive setting and always use the current batch statistics.

\begin{table}[t]
\centering
\caption{Comparison of accuracy (\%) with 95\% confidence intervals (10 iterations) with bitwidth-dedicated and bitwidth-adaptive QAT methods. $\dagger$ denotes results from \cite{CuiLY0CKX20}. $\ddag$ denotes results from a non-differentiated binarization function. FP stands for 32-bit Full-Precision. '-' denotes results not provided.}
\label{table:comparison_yu}
\resizebox{\textwidth}{!}{%
\begin{tabular}{|c|ccc|ccc|ccc|}
\toprule
\multirow{2}{*}{($b_w, b_a$)} & \multicolumn{3}{c|}{CIFAR-10, MobileNet-v2}        & \multicolumn{3}{c|}{CIFAR-10, Pre-activation ResNet-20} & \multicolumn{3}{c|}{SVHN, 8-layer CNN}                         \\ \cline{2-9} 
                              & \multicolumn{1}{c|}{QAT} & \multicolumn{1}{c|}{AdaBits} & MEBQAT                    & \multicolumn{1}{c|}{QAT}   & \multicolumn{1}{c|}{ApDNN}    & MEBQAT                             & \multicolumn{1}{c|}{QAT} & \multicolumn{1}{c|}{ApDNN} & MEBQAT \\ \hline\hline
(1, 1)                        & -                        & -                            & -                         & $92.28{\;}({\pm}0.116)$    & $92.15^{\ddag}$               & $91.32{\;}({\pm}0.202)$     & $97.27{\;}({\pm}0.025)$         & $88.21^{\ddag}$           & $96.60{\;}({\pm}0.060)$  \\
(2, 2)                        & $84.40{\;}({\pm}0.691)$  & $58.98^{\dagger}$            & $78.50{\;}({\pm}0.544)$   & $92.72{\;}({\pm}0.146)$    & 93.97                         & $92.52{\;}({\pm}0.151)$     & $97.51{\;}({\pm}0.043)$         & 94.94                     & $97.25{\;}({\pm}0.052)$  \\
(3, 3)                        & $90.08{\;}({\pm}0.233)$  & $79.30^{\dagger}$            & $88.04{\;}({\pm}0.255)$   & $92.61{\;}({\pm}0.066)$    & -                             & $92.65{\;}({\pm}0.225)$     & $97.57{\;}({\pm}0.024)$         & -                         & $97.58{\;}({\pm}0.041)$  \\
(4, 4)                        & $90.44{\;}({\pm}0.152)$  & $91.84^{\dagger}$            & $89.30{\;}({\pm}0.336)$   & $92.69{\;}({\pm}0.195)$    & 93.95                         & $92.77{\;}({\pm}0.157)$     & $97.44{\;}({\pm}0.068)$         & 96.19                     & $97.62{\;}({\pm}0.043)$  \\
(5, 5)                        & $90.83{\;}({\pm}0.193)$  & -                            & $89.58{\;}({\pm}0.243)$   & $92.64{\;}({\pm}0.117)$    & -                             & $92.80{\;}({\pm}0.179)$     & $97.53{\;}({\pm}0.028)$         & -                         & $97.64{\;}({\pm}0.050)$  \\
(6, 6)                        & $91.10{\;}({\pm}0.146)$  & -                            & $89.46{\;}({\pm}0.275)$   & $92.66{\;}({\pm}0.120)$    & -                             & $92.83{\;}({\pm}0.188)$     & $97.50{\;}({\pm}0.032)$         & -                             & $97.63{\;}({\pm}0.056)$  \\
(7, 7)                        & $91.06{\;}({\pm}0.138)$  & -                            & $89.48{\;}({\pm}0.303)$   & $92.65{\;}({\pm}0.110)$    & -                             & $92.79{\;}({\pm}0.171)$     & $97.56{\;}({\pm}0.034)$         & -                             & $97.64{\;}({\pm}0.043)$  \\
(8, 8)                        & $91.20{\;}({\pm}0.171)$  & -                            & $89.36{\;}({\pm}0.243)$   & $92.57{\;}({\pm}0.124)$    & 93.80                         & $92.89{\;}({\pm}0.147)$     & $97.52{\;}({\pm}0.055)$         & 96.22                     & $97.63{\;}({\pm}0.047)$  \\
(16, 16)                      & $91.19{\;}({\pm}0.145)$  & -                            & $89.53{\;}({\pm}0.209)$   & $92.67{\;}({\pm}0.192)$    & -                             & $92.75{\;}({\pm}0.190)$     & $97.51{\;}({\pm}0.042)$         & -                             & $97.65{\;}({\pm}0.056)$  \\
(FP, FP)                      & $93.00{\;}({\pm}0.221)$  & -                            & $89.24{\;}({\pm}0.253)$   & $93.92{\;}({\pm}0.107)$    & 93.98                         & $92.90{\;}({\pm}0.133)$     & $97.67{\;}({\pm}0.079)$         & 96.29                     & $97.40{\;}({\pm}0.043)$  \\ \hline
\end{tabular}}
\end{table}

\subsubsection{Performance of \proposal.} 
Table 
\ref{table:comparison_yu} shows the (meta-)test accuracy after (meta-)trained by QAT/bitwidth-adaptive QAT/\proposal in multiple model architectures and datasets. Here, $b_w, b_a$ are bitwidths used during testing. For each bitwidth setting, accuracy is averaged over one test epoch. 
Results of vanilla QAT come from individually trained models dedicated to a single bitwidth. All other results come from a single adaptable model, albeit with some prior work containing bitwidth-dedicated parts. Results show that \proposal achieves performance comparable to or better than the existing methods.


\begin{table}[t]
\caption{Comparison of training computation and storage costs.}
\centering
\resizebox{0.7 \columnwidth}{!}{%
\label{table:comp_cost}
\begin{tabular}{l|cc}
\hline
Methods         & Training computation cost     & Storage cost \\ \hline
Dedicated QAT             & 1 backprop per update         & $T\Theta$ \\
AdaBits/ApDNN   & $T$ backprops per update      & $(1-\zeta)\Theta + T\zeta\Theta$ \\
MEBQAT          & $M$ backprops per update      & $\Theta$ \\ \hline
\end{tabular}}
\end{table}

We also tackle the limitations of prior bitwidth-adaptive QAT methods in scalability to the number of target bitwidths. Table \ref{table:comp_cost} shows an overview of training and storage costs of various methods when compared with \proposal. Here, $T({\simeq}{|\mathcal{T}_b|}^2)$ represents the number of (test) bitwidths, $\Theta$ denotes the total model size, and $\zeta$ indicates the ratio of batch normalization layers respective to the entire model. Because \proposal is a meta-learning alternative to bitwidth adaptive learning, our method exhibits fast adaptation, requiring only a few train steps $M$. In evaluation scenarios, $1<M(=4)\ll T(=73 \;\text{or}\; 75)$, showing that \proposal is up to 18 times more cost-efficient than other methods since it trains a single model with a single batch normalization layer for all different tasks. Note that computation costs are the same for all non-few-shot methods during testing since inference directly follows quantization. In other words, \proposal requires zero additional training during inference. Thus, \proposal exhibits much more training efficiency than other adaptive methods in non-few-shot scenarios.

\begin{table}[t]
\centering
\caption{Comparison of accuracy (\%) to vanilla FOMAML and FOMAML + QAT, using 5-layer CNN in \cite{FinnAL17}.}
\label{table:comparison_fomaml}
\resizebox{0.9\textwidth}{!}{%
\begin{tabular}{|c|ccc|ccc|}
\toprule
\multirow{2}{*}{$(b_w, b_a)$} & \multicolumn{3}{c|}{Omniglot 20-way 1-shot, 5-layer CNN}                    & \multicolumn{3}{c|}{Omniglot 20-way 5-shot, 5-layer CNN}                    \\ \cline{2-7} 
                              & \multicolumn{1}{c|}{FOMAML} & \multicolumn{1}{c|}{FOMAML+QAT} & MEBQAT-MAML & \multicolumn{1}{c|}{FOMAML} & \multicolumn{1}{c|}{FOMAML+QAT} & MEBQAT-MAML \\ \hline\hline
(2, 2)                        & 25.97                       & 62.09                           & 89.57       & 35.24                       & 84.03                           & 96.94       \\
(3, 3)                        & 75.29                       & 65.24                           & 91.46       & 83.29                       & 83.29                           & 97.58       \\
(4, 4)                        & 84.43                       & 63.84                           & 91.62       & 88.19                       & 84.73                           & 97.61       \\
(5, 5)                        & 89.51                       & 67.35                           & 91.65       & 93.28                       & 97.78                           & 97.61       \\
(6, 6)                        & 91.47                       & 92.95                           & 91.66       & 96.43                       & 97.53                           & 97.61       \\
(7, 7)                        & 90.94                       & 92.40                           & 91.65       & 96.61                       & 97.41                           & 97.61       \\
(8, 8)                        & 91.92                       & 93.00                           & 91.66       & 97.20                       & 97.86                           & 97.61       \\
(16, 16)                      & 93.13                       & 92.82                           & 91.65       & 97.47                       & 97.41                           & 97.69       \\
(FP, FP)                      & 93.12                       & 93.12                           & 92.39       & 97.48                       & 97.48                           & 97.88       \\ \hline
                              & \multicolumn{3}{c|}{MiniImageNet 5-way 1-shot, 5-layer CNN}                 & \multicolumn{3}{c|}{MiniImageNet 5-way 5-shot, 5-layer CNN}                 \\ \hline\hline
(2, 2)                        & 34.96                       & 42.35                           & 46.00       & 47.49                       & 62.25                           & 61.65       \\
(3, 3)                        & 43.89                       & 42.11                           & 47.45       & 59.02                       & 63.16                           & 63.82       \\
(4, 4)                        & 47.14                       & 48.75                           & 47.56       & 63.40                       & 64.76                           & 63.54       \\
(5, 5)                        & 48.19                       & 47.07                           & 47.45       & 64.09                       & 65.54                           & 63.69       \\
(6, 6)                        & 48.56                       & 48.66                           & 47.46       & 64.31                       & 64.09                           & 63.67       \\
(7, 7)                        & 48.62                       & 48.10                           & 47.43       & 64.41                       & 64.57                           & 63.72       \\
(8, 8)                        & 48.60                       & 48.26                           & 47.43       & 64.53                       & 64.65                           & 63.70       \\
(16, 16)                      & 48.65                       & 48.17                           & 47.36       & 64.48                       & 65.17                           & 63.81       \\
(FP, FP)                      & 48.66                       & 48.66                           & 47.68       & 64.51                       & 64.51                           & 64.28       \\ \hline
\end{tabular}}
\end{table}

\begin{table}[t]
\centering
\caption{Comparison of accuracy (\%) to vanilla PN and PN + QAT, using 4-layer CNN in \cite{SnellSZ17}.}
\label{table:comparison_protonet}
\resizebox{0.9\textwidth}{!}{%
\begin{tabular}{|c|ccc|ccc|}
\toprule
\multirow{2}{*}{$(b_w, b_a)$} & \multicolumn{3}{c|}{Omniglot 20-way 1-shot, 4-layer CNN}          & \multicolumn{3}{c|}{Omniglot 20-way 5-shot, 4-layer CNN}          \\ \cline{2-7} 
                              & \multicolumn{1}{c|}{PN} & \multicolumn{1}{c|}{PN+QAT} & MEBQAT-PN & \multicolumn{1}{c|}{PN} & \multicolumn{1}{c|}{PN+QAT} & MEBQAT-PN \\ \hline\hline
(2, 2)                        & 31.58                   & 95.46                       & 94.87     & 50.86                   & 98.71                       & 98.32     \\
(3, 3)                        & 81.21                   & 95.90                       & 95.55     & 93.87                   & 98.77                       & 98.56     \\
(4, 4)                        & 93.73                   & 95.97                       & 95.60     & 98.37                   & 98.76                       & 98.58     \\
(5, 5)                        & 95.40                   & 95.95                       & 95.60     & 98.77                   & 98.61                       & 98.58     \\
(6, 6)                        & 95.83                   & 95.82                       & 95.60     & 98.83                   & 98.65                       & 98.58     \\
(7, 7)                        & 95.84                   & 95.95                       & 95.60     & 98.87                   & 98.62                       & 98.58     \\
(8, 8)                        & 95.88                   & 95.97                       & 95.60     & 98.88                   & 98.85                       & 98.58     \\
(16, 16)                      & 96.89                   & 95.55                       & 95.60     & 98.89                   & 98.93                       & 98.58     \\
(FP, FP)                      & 95.88                   & 95.88                       & 96.06     & 98.89                   & 98.89                       & 98.70     \\ \hline
                              & \multicolumn{3}{c|}{MiniImageNet 5-way 1-shot, 4-layer CNN}       & \multicolumn{3}{c|}{MiniImageNet 5-way 5-shot, 4-layer CNN}       \\ \hline\hline
(2, 2)                        & 26.29                   & 50.06                       & 47.66     & 30.64                   & 67.45                       & 65.34     \\
(3, 3)                        & 37.51                   & 50.16                       & 48.57     & 46.74                   & 67.71                       & 66.16     \\
(4, 4)                        & 45.59                   & 50.38                       & 48.38     & 60.52                   & 67.35                       & 66.22     \\
(5, 5)                        & 48.33                   & 50.18                       & 48.54     & 64.75                   & 65.95                       & 66.16     \\
(6, 6)                        & 49.77                   & 50.01                       & 48.55     & 65.81                   & 65.63                       & 66.19     \\
(7, 7)                        & 49.52                   & 49.90                       & 48.55     & 65.68                   & 66.06                       & 66.19     \\
(8, 8)                        & 49.29                   & 48.35                       & 48.55     & 65.90                   & 65.86                       & 66.19     \\
(16, 16)                      & 49.75                   & 47.86                       & 48.55     & 65.94                   & 66.39                       & 66.19     \\
(FP, FP)                      & 49.61                   & 49.61                       & 48.33     & 65.82                   & 65.82                       & 66.03     \\ \hline
\end{tabular}}
\end{table}

\subsection{Experiments on the Bitwidth-Class Joint Adaptation Scenario}

To the best of our knowledge, there is no prior work on multi-bit quantization in a few-shot context. Therefore, we compare \proposal-MAML and \proposal-PN to two types of compared schemes: (1) vanilla meta-learning without quantization-awareness and (2) meta-learning combined with bitwidth-dedicated QAT. In (2), by using fake-quantized $b$-bit models in conventional meta-learning operations, the model shows solid adaptable performance in $b$-bits. Just as in section \eqref{section:experiments_traditional}, we conduct experiments with much more various bitwidth candidates than existing QAT-based methods.  


%
When using the MAML framework, there are 16/4 inner-loop tasks using Omniglot/MiniImageNet, respectively. In an inner-loop, the 5-layer CNN in \cite{FinnAL17} is updated by a SGD optimizer with learning rate $10^{-1}$/$10^{-2}$ at 5 times with a support set. In an outer-loop, the base model is trained by Adam optimizer with learning rate $10^{-4}$. In the meta-testing phase, fine-tuning occurs in 5/10 times, with an optimizer same as inner-loop optimizer in the previous phase.
When using the PN framework, a model is optimized by Adam with learning rate $10^{-3}$. We use Euclidean distance as a metric for classification. \proposal-PN has 4 inner-loop tasks per outer-loop.


\subsubsection{Performance of \proposal-MAML and \proposal-PN.} Table \ref{table:comparison_fomaml} shows the meta-testing accuracy after meta-trained by FOMAML, FOMAML + QAT and \proposal-MAML. For each bitwidth setting, accuracy is averaged over 600 different sets of $N$ target classes unseen in the previous phase. It is noteworthy that in some cases, \proposal-MAML exceeds the postulated upper bound of accuracy. In other words, although we hypothesized applying bitwidth-dedicated QAT directly to train individual models would have the highest accuracy, we found that in some cases, \proposal-MAML achieves performance exceeding the baseline.

Table \ref{table:comparison_protonet} shows the meta-testing accuracy after meta-trained by PN, PN + QAT and \proposal-PN. For each bitwidth setting, accuracy is averaged over 600 different sets of $N$ target classes unseen in the previous phase. The results prove that \proposal is also compliant to metric-based meta-learning such that the base model can fit into any target bitwidth as well as target classes without fine-tuning in the test side. 

\section{Discussion}
Although this paper focuses on quantizing the entire model into a single bitwidth, and increasingly growing area of research focuses on quantizing each layer or block of the model into different optimal bitwidths. When \proposal is directly applied to this mixed-precision setting, this might require many diverse tasks, which poses heavier computational burdens both during training and when finding an optimal bitwidth for each platform during inference. Development of an efficient meta-learning method for both adaptive- and mixed-precision quantization would be an interesting future work.  

A limitation of our current experiments comes from the fact that in our method, QAT does not consist solely of integer-arithmetic-only operations. Moreover, \proposal-MAML stipulates fine-tuning at meta-testing phase for adaptation, where the gradient descent during this process is mostly done in full precision. In this case, future work can include applying integer-only methods such as in HAWQ-v3 \cite{YaoDZGYTWHWMK21} as a quantization(-aware training) method to further test the feasibility of our method. We can also proceed to use COTS edge devices such as a Coral development board to evaluate the applicability of our method.

Increasing the performance of adaptability of our work is another future work. This is especially true since FOMAML and Prototypical Networks are methods that have been tried and tested for several years. Using other sophisticated meta-learning methods can improve the adaptability performance of our model or reduce the computational complexity of fine-tuning our model at a resource-constrained device. 

\section{Conclusion}
To the best of our knowledge, this paper is the first to attempt training a model with meta-learning which can be independently quantized to any arbitrary bitwidth at runtime. To this end, we investigate the possibility of incorporating bitwidths as an adaptable meta-task, and propose a method by which the model can be trained to adapt into any bitwidth, as well as any target classes in a supervised-learning context. Through experimentation, we found that our proposed method achieves performance greater than or equal to existing work on adaptable bitwidths, showing that incorporating meta-learning could become a viable alternative. We also found that our method is robust to a few-shot learning context, showing better performance than models trained with dedicated meta-learning techniques and quantized using PTQ or QAT. Thus, we demonstrate that \proposal can potentially open up an interesting new avenue of research in the field of bitwidth-adaptive QAT.


\section*{Acknowledgements}
This research was supported in part by the MSIT (Ministry of Science and ICT), Korea, under the ITRC (Information Technology Research Center) support program (IITP-2021-0-02048) supervised by the IITP (Institute of Information \& Communications Technology Planning \& Evaluation), the National Research Foundation of Korea (NRF) grant (No. 2020R1A2C2101815), and Samsung Research Funding \& Incubation Center of Samsung Electronics under Project No. SRFC-TD2003-01. 


\clearpage
%
%
\bibliographystyle{splncs04}
\bibliography{egbib}

\begin{thebibliography}{10}
\providecommand{\url}[1]{\texttt{#1}}
\providecommand{\urlprefix}{URL }
\providecommand{\doi}[1]{https://doi.org/#1}

\bibitem{BaiCHS21}
Bai, H., Cao, M., Huang, P., Shan, J.: Batchquant: Quantized-for-all
  architecture search with robust quantizer. CoRR  \textbf{abs/2105.08952}
  (2021)

\bibitem{BulatT21}
Bulat, A., Tzimiropoulos, G.: Bit-mixer: Mixed-precision networks with runtime
  bit-width selection. In: Proceedings of the IEEE/CVF International Conference
  on Computer Vision (ICCV). pp. 5188--5197 (2021)

\bibitem{CaiYDGMK20}
Cai, Y., Yao, Z., Dong, Z., Gholami, A., Mahoney, M.W., Keutzer, K.: Zeroq: A
  novel zero shot quantization framework. In: Proceedings of the IEEE
  conference on Computer Vision and Pattern Recognition (CVPR). pp.
  13166--13175 (2020)

\bibitem{cho2021camera}
Cho, H., Cho, Y., Yu, J., Kim, J.: Camera distortion-aware 3d human pose
  estimation in video with optimization-based meta-learning. In: Proceedings of
  the IEEE/CVF International Conference on Computer Vision. pp. 11169--11178
  (2021)

\bibitem{ChoukrounKYK19}
Choukroun, Y., Kravchik, E., Yang, F., Kisilev, P.: Low-bit quantization of
  neural networks for efficient inference. In: Proceedings of the IEEE/CVF
  International conference on Computer Vision (ICCV) Workshops. pp. 3009--3018
  (2019)

\bibitem{CuiLY0CKX20}
Cui, Y., Liu, Z., Yao, W., Li, Q., Chan, A.B., Kuo, T., Xue, C.J.: Fully nested
  neural network for adaptive compression and quantization. In: Proceedings of
  the International Joint conference on Artificial Intelligence (IJCAI). pp.
  2080--2087 (2020)

\bibitem{EsserMBAM20}
Esser, S.K., McKinstry, J.L., Bablani, D., Appuswamy, R., Modha, D.S.: Learned
  step size quantization. In: Proceedings of the International conference on
  Learning Representations (ICLR) (2020)

\bibitem{FangSATGH20}
Fang, J., Shafiee, A., Abdel{-}Aziz, H., Thorsley, D., Georgiadis, G., Hassoun,
  J.: Post-training piecewise linear quantization for deep neural networks. In:
  Proceedings of the European Conference on Computer Vision (ECCV). vol. 12347,
  pp. 69--86 (2020)

\bibitem{FinnAL17}
Finn, C., Abbeel, P., Levine, S.: Model-agnostic meta-learning for fast
  adaptation of deep networks. In: Proceedings of the International conference
  on Machine Learning (ICML). vol.~70, pp. 1126--1135 (2017)

\bibitem{FinnXL18}
Finn, C., Xu, K., Levine, S.: Probabilistic model-agnostic meta-learning. In:
  Proceedings of the advances in Neural Information Processing Systems
  (NeurIPS). pp. 9537--9548 (2018)

\bibitem{GongLJLHLYY19}
Gong, R., Liu, X., Jiang, S., Li, T., Hu, P., Lin, J., Yu, F., Yan, J.:
  Differentiable soft quantization: Bridging full-precision and low-bit neural
  networks. In: Proceedings of the IEEE/CVF International conference on
  Computer Vision (ICCV). pp. 4851--4860 (2019)

\bibitem{gong2019metasense}
Gong, T., Kim, Y., Shin, J., Lee, S.J.: Metasense: few-shot adaptation to
  untrained conditions in deep mobile sensing. In: Proceedings of the 17th
  Conference on Embedded Networked Sensor Systems. pp. 110--123 (2019)

\bibitem{GuoZMHLWS20}
Guo, Z., Zhang, X., Mu, H., Heng, W., Liu, Z., Wei, Y., Sun, J.: Single path
  one-shot neural architecture search with uniform sampling. In: Proceedings of
  the European Conference on Computer Vision (ECCV). vol. 12361, pp. 544--560
  (2020)

\bibitem{HintonVD15}
Hinton, G.E., Vinyals, O., Dean, J.: Distilling the knowledge in a neural
  network (2015)

\bibitem{HubaraNHBS21}
Hubara, I., Nahshan, Y., Hanani, Y., Banner, R., Soudry, D.: Accurate post
  training quantization with small calibration sets. In: Proceedings of the
  International conference on Machine Learning (ICML). vol.~139, pp. 4466--4475
  (2021)

\bibitem{JacobKCZTHAK18}
Jacob, B., Kligys, S., Chen, B., Zhu, M., Tang, M., Howard, A.G., Adam, H.,
  Kalenichenko, D.: Quantization and training of neural networks for efficient
  integer-arithmetic-only inference. In: Proceedings of the IEEE conference on
  Computer Vision and Pattern Recognition (CVPR). pp. 2704--2713 (2018)

\bibitem{JinYL20}
Jin, Q., Yang, L., Liao, Z.: Adabits: Neural network quantization with adaptive
  bit-widths. In: Proceedings of the IEEE/CVF conference on Computer Vision and
  Pattern Recognition (CVPR). pp. 2143--2153 (2020)

\bibitem{JinYLQ20}
Jin, Q., Yang, L., Liao, Z., Qian, X.: Neural network quantization with
  scale-adjusted training. In: Proceedings of the British Machine Vision
  Conference (BMVC) (2020)

\bibitem{JungSLSHKHC19}
Jung, S., Son, C., Lee, S., Son, J., Han, J., Kwak, Y., Hwang, S.J., Choi, C.:
  Learning to quantize deep networks by optimizing quantization intervals with
  task loss. In: Proceedings of the IEEE conference on Computer Vision and
  Pattern Recognition (CVPR). pp. 4350--4359 (2019)

\bibitem{nvidia_quantization}
Lee, J.: Fast int8 inference for autonomous vehicles with tensorrt 3 (Dec
  2017),
  \url{https://developer.nvidia.com/blog/int8-inference-autonomous-vehicles-tensorrt/}

\bibitem{LiGTYHZYWG21}
Li, Y., Gong, R., Tan, X., Yang, Y., Hu, P., Zhang, Q., Yu, F., Wang, W., Gu,
  S.: Brecq: Pushing the limit of post-training quantization by block
  reconstruction. In: Proceedings of the International conference on Learning
  Representations (ICLR) (2021)

\bibitem{LiZCL17}
Li, Z., Zhou, F., Chen, F., Li, H.: Meta-sgd: Learning to learn quickly for few
  shot learning. CoRR  \textbf{abs/1707.09835} (2017)

\bibitem{LiangGWSZ21}
Liang, T., Glossner, J., Wang, L., Shi, S., Zhang, X.: Pruning and quantization
  for deep neural network acceleration: A survey. Neurocomputing  \textbf{461},
   370--403 (2021)

\bibitem{Mallat89}
Mallat, S.: A theory for multiresolution signal decomposition: the wavelet
  representation. IEEE Transactions on Pattern Analysis and Machine
  Intelligence (TPAMI)  \textbf{11}(7),  674--693 (1989)

\bibitem{MathieuHL13}
Mathieu, M., Henaff, M., LeCun, Y.: Fast training of convolutional networks
  through ffts. In: Proceedings of the International conference on Learning
  Representations (ICLR) (2014)

\bibitem{RajeswaranFKL19}
Rajeswaran, A., Finn, C., Kakade, S.M., Levine, S.: Meta-learning with implicit
  gradients. In: Proceedings of the advances in Neural Information Processing
  Systems (NeurIPS). pp. 113--124 (2019)

\bibitem{RaviB19}
Ravi, S., Beatson, A.: Amortized bayesian meta-learning. In: Proceedings of the
  International conference on Learning Representations (ICLR) (2019)

\bibitem{SainathKSAR13}
Sainath, T.N., Kingsbury, B., Sindhwani, V., Arisoy, E., Ramabhadran, B.:
  Low-rank matrix factorization for deep neural network training with
  high-dimensional output targets. In: Proceedings of the International
  conference on Acoustics, Speech, and Signal Processing (ICASSP). pp.
  6655--6659 (2013)

\bibitem{ShenLLLSYO20}
Shen, M., Liang, F., Li, C., Lin, C., Sun, M., Yan, J., Ouyang, W.: Once
  quantization-aware training: High performance extremely low-bit architecture
  search. CoRR  \textbf{abs/2010.04354} (2020)

\bibitem{SnellSZ17}
Snell, J., Swersky, K., Zemel, R.S.: Prototypical networks for few-shot
  learning. In: Proceedings of the advances in Neural Information Processing
  Systems (NeurIPS). pp. 4077--4087 (2017)

\bibitem{SunLRHLJL21}
Sun, Q., Li, X., Ren, Y., Huang, Z., Liu, X., Jiao, L., Liu, F.: One model for
  all quantization: A quantized network supporting hot-swap bit-width
  adjustment (2021)

\bibitem{SungZXHY17}
Sung, F., Zhang, L., Xiang, T., Hospedales, T.M., Yang, Y.: Learning to learn:
  Meta-critic networks for sample efficient learning. CoRR
  \textbf{abs/1706.09529} (2017)

\bibitem{SuzukiAMHIWHYN20}
Suzuki, T., Abe, H., Murata, T., Horiuchi, S., Ito, K., Wachi, T., Hirai, S.,
  Yukishima, M., Nishimura, T.: Spectral pruning: Compressing deep neural
  networks via spectral analysis and its generalization error. In: Proceedings
  of the International Joint conference on Artificial Intelligence (IJCAI). pp.
  2839--2846 (2020)

\bibitem{SwaminathanGKA20}
Swaminathan, S., Garg, D., Kannan, R., Andres, F.: Sparse low rank
  factorization for deep neural network compression. Neurocomputing
  \textbf{398},  185--196 (2020)

\bibitem{TriantafillouZDLEXGGSML20}
Triantafillou, E., Zhu, T., Dumoulin, V., Lamblin, P., Evci, U., Xu, K.,
  Goroshin, R., Gelada, C., Swersky, K., Manzagol, P., Larochelle, H.:
  Meta-dataset: A dataset of datasets for learning to learn from few examples.
  In: Proceedings of the International conference on Learning Representations
  (ICLR) (2020)

\bibitem{WangLLLH19}
Wang, K., Liu, Z., Lin, Y., Lin, J., Han, S.: Haq: Hardware-aware automated
  quantization with mixed precision. In: Proceedings of the IEEE conference on
  Computer Vision and Pattern Recognition (CVPR). pp. 8612--8620 (2019)

\bibitem{WangWCLLWLH20}
Wang, T., Wang, K., Cai, H., Lin, J., Liu, Z., Wang, H., Lin, Y., Han, S.: Apq:
  Joint search for network architecture, pruning and quantization policy. In:
  Proceedings of the IEEE/CVF conference on Computer Vision and Pattern
  Recognition (CVPR). pp. 2075--2084 (2020)

\bibitem{WuWZTVK18}
Wu, B., Wang, Y., Zhang, P., Tian, Y., Vajda, P., Keutzer, K.: Mixed precision
  quantization of convnets via differentiable neural architecture search. CoRR
  \textbf{abs/1812.00090} (2018)

\bibitem{YaoDZGYTWHWMK21}
Yao, Z., Dong, Z., Zheng, Z., Gholami, A., Yu, J., Tan, E., Wang, L., Huang,
  Q., Wang, Y., Mahoney, M.W., Keutzer, K.: {HAWQ-V3:} dyadic neural network
  quantization. In: Proceedings of the International Conference on Machine
  Learning (ICML). vol.~139, pp. 11875--11886 (2021)

\bibitem{YoonKDKBA18}
Yoon, J., Kim, T., Dia, O., Kim, S., Bengio, Y., Ahn, S.: Bayesian
  model-agnostic meta-learning. In: Proceedings of the advances in Neural
  Information Processing Systems (NeurIPS). pp. 7343--7353 (2018)

\bibitem{YuLSH021}
Yu, H., Li, H., Shi, H., Huang, T.S., Hua, G.: Any-precision deep neural
  networks. In: Proceedings of the AAAI conference on Artificial Intelligence
  (AAAI). pp. 10763--10771 (2021)

\bibitem{YuLCLMHGLD18}
Yu, R., Li, A., Chen, C., Lai, J., Morariu, V.I., Han, X., Gao, M., Lin, C.,
  Davis, L.S.: Nisp: Pruning networks using neuron importance score
  propagation. In: Proceedings of the IEEE conference on Computer Vision and
  Pattern Recognition (CVPR). pp. 9194--9203 (2018)

\bibitem{ZhangYYH18}
Zhang, D., Yang, J., Ye, D., Hua, G.: Lq-nets: Learned quantization for highly
  accurate and compact deep neural networks. In: Proceedings of the European
  Conference on Computer Vision (ECCV). vol. 11212, pp. 373--390 (2018)

\bibitem{ZhouZYFXH21}
Zhou, P., Zou, Y., Yuan, X., Feng, J., Xiong, C., Hoi, S.C.H.: Task similarity
  aware meta learning: theory-inspired improvement on maml. In: Proceedings of
  the conference on Uncertainty in Artificial Intelligence (UAI). vol.~161, pp.
  23--33 (2021)

\bibitem{ZhouNZWWZ16}
Zhou, S., Ni, Z., Zhou, X., Wen, H., Wu, Y., Zou, Y.: Dorefa-net: Training low
  bitwidth convolutional neural networks with low bitwidth gradients. CoRR
  \textbf{abs/1606.06160} (2016)

\end{thebibliography}
\end{document}